\documentclass[letterpaper, 10 pt, conference]{ieeeconf}  
\usepackage{lmodern}

\IEEEoverridecommandlockouts                              

\usepackage{amssymb}  

\usepackage[fleqn]{amsmath}
\usepackage{graphicx}
\usepackage{color}
\usepackage{amsmath} 
\usepackage[table,xcdraw]{xcolor}
\usepackage{caption}
\usepackage{booktabs}
\usepackage{threeparttable}
\usepackage{amsfonts}
\usepackage{diagbox}
\usepackage{slashbox}

\title{\LARGE \bf MagicGel: A Novel Visual-Based Tactile Sensor Design with Magnetic Gel}

\author{ Jianhua Shan$^{1}$, Jie Zhao$^{1,2}$, Jiangduo Liu$^{3}$, Xiangbo Wang$^{4}$, Ziwei Xia$^{5}$, Guangyuan Xu$^{2}$ and Bin Fang$^{2}$ 
\thanks{J Shan and J Zhao are co-first authors of the article.}
\thanks{Corresponding author: Bin Fang.}
\thanks{$^{1}$School of Mechanical Engineering, Anhui University of Technology, Ma'anshan 243032, China. {\tt\small2931@ahut.edu.cn. jha06039@gmail.com.} }%
\thanks{$^{2}$School of Artificial Intelligence, Beijing University of Posts and Telecommunications, 
Beijing 100876.xuguangyuan@bupt.edu.cn,
 {\tt\small fangbin1120@bupt.edu.cn}.}%
\thanks{$^{3}$School of Intelligence Science and Technology, University of Science and Technology Beijing, Beijing 10083, China. 18438873695@163.com.}%
\thanks{$^{4}$College of Quality and Technical Supervision, Hebei University, Baoding 071002, China  15369902997@163.com.}%
\thanks{$^{5}$ School of Engineering and Technology, China University of Geosciences (Beijing), Beijing 10083 China. xzw@email.cugb.edu.cn.}%
}

\begin{document}

\maketitle
\thispagestyle{empty}
\pagestyle{empty}

\begin{abstract}

Force estimation is the core indicator for evaluating the performance of tactile sensors, and it is also the key technical path to achieve precise force feedback mechanisms. This study proposes a design method for a visual tactile sensor (VBTS) that integrates a magnetic perception mechanism, and develops a new tactile sensor called MagicGel. The sensor uses strong magnetic particles as markers and captures magnetic field changes in real time through Hall sensors. On this basis, MagicGel achieves the coordinated optimization of multimodal perception capabilities: it not only has fast response characteristics, but also can perceive non-contact status information of home electronic products. Specifically, MagicGel simultaneously analyzes the visual characteristics of magnetic particles and the multimodal data of changes in magnetic field intensity, ultimately improving force estimation capabilities.

\end{abstract}


\section{INTRODUCTION}

With the rapid advancement of tactile sensor technology, its crucial role in robotics, automation systems, and human-computer interaction has become increasingly evident. Tactile sensors enhance a robot's ability to perceive its environment, equipping the robot with more precise and intelligent operational capabilities. 
In the field of flexible operation and human-computer interaction, accurate tactile perception is the key to realizing core functions such as bionic grasping and force-controlled interaction. Traditional tactile sensors are mostly based on piezoresistance, capacitance or piezoelectric principles, which can achieve quantitative force perception. However, they have significant limitations in spatial resolution, dynamic response range and force estimation accuracy.

VBTS (Vision-Based Tactile Sensor) is an advanced sensor that employ a camera to capture deformation data from the tactile surface of the sensor, as well as the displacement and deformation of internal markers. Since the emergence of Gelsight \cite{johnson2009retrographic} and Tactip\cite{chorley2009development} sensors in 2009, VBTS has been extensively employed in various tactile perception applications due to their high resolution and robust resistance to interference. These applications encompass operational tasks performed under occlusion and in dark, narrow spaces. By harnessing the capabilities of high-resolution tactile sensors, dexterous robotic hands can enhance their perceptual abilities, such as texture recognition \cite{lu2024innovative} and force measurement \cite{zhang2022learning, sun2022soft, funk2023high, li2024biotactip}. This enhancement enables dexterous hands to execute manipulation tasks that require sensitive tactile feedback \cite{kim2022uvtac, she2021cable}. By fully utilizing feedback information during the operational process, high-resolution tactile sensors can detect subtle changes while shaking a bottle, thereby determining the type of substance contained within. For example, these sensors can be used to identify solid \cite{guo2023estimating} and liquid \cite{huang2022understanding} objects in unknown bottles. Additionally, VBTSs can assist dexterous hands in optimizing tool usage \cite{song2019sensing}. Compared to other types of tactile sensors, VBTSs demonstrate superior robustness. For instance, when a dexterous hand grasps a knife \cite{yamaguchi2016combining} or a screwdriver \cite{dong2019maintaining}, it can more effectively adjust the object in hand based on the feedback received \cite{wang2020swingbot}. Crucially, the foundation for executing these operations lies in achieving a high-precision estimation of the surface contact force.

\begin{figure}[t]
\centering 
\includegraphics[width=0.5\textwidth]{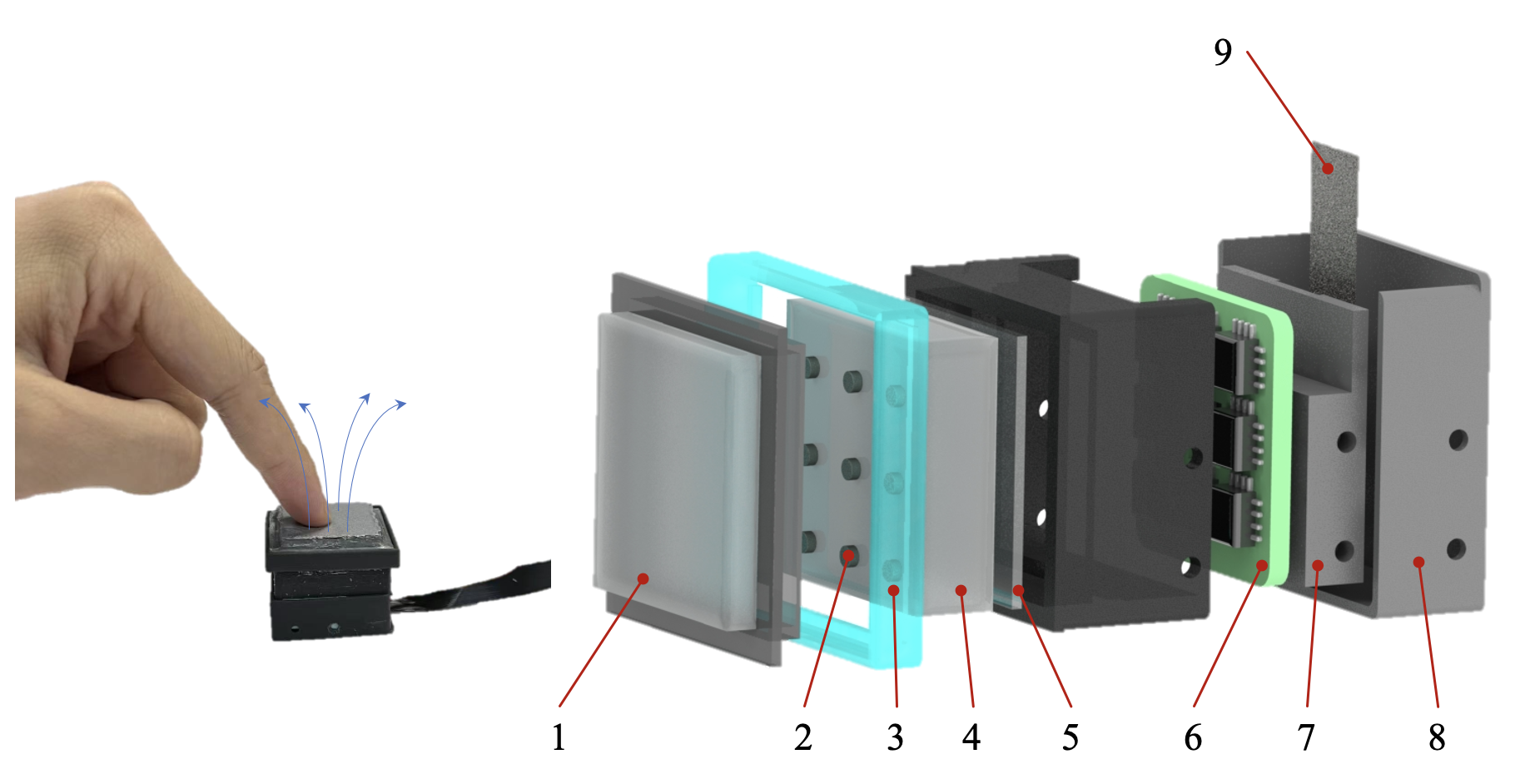} 
\caption{MagicGe structure diagram. \textbf{(1)} Elastomer 1. \textbf{(2)} Magnetic particle marking points. \textbf{(3)} Light source \textbf{(4)} Elastomer 2. \textbf{(5)} Support plate. \textbf{(6)} Hall sensor circuit board. \textbf{(7)} Connectors. \textbf{(8)} Base. \textbf{(9)} Camera.
} 
\label{fig:1} 
\end{figure}

The manufacturing process and processing algorithms\cite{guo2024aerial} of VBTS continue to evolve, demonstrating a trend of continuous improvement. However, the response speed of its sensors is still primarily limited by the data acquisition frequency of the camera and the speed of backend image processing, which to some extent constrains the rapid response capability of the visual-tactile sensors to complex environments.

In terms of perceptual performance, magnetic tactile sensors hold a significant advantage over visual-tactile sensors, mainly manifested in their high response rate and superior adaptability. Magnetic tactile sensors rapidly detect changes in magnetic fields to sense contact force and area. These sensors are primarily categorized into Hall effect type\cite{ci1} and Faraday electromagnetic induction effect type\cite{ci2}. Hall effect magnetic tactile sensors can effectively recognize continuous force contact signals, so this paper adopts the strong magnetic particles and Hall sensors.

To compensate for the deficiencies in image marking information reception of VBTS, this study explores a method that integrates magnetic mechanisms with visual-tactile sensors. By introducing magnetic field information to supplement image information, the dimensional expansion of the 2D displacement field is achieved. Main contributions:

1. A new tactile sensor is designed by Vision-magnetic fusion method. Using magnetic particles as markers for VBTS, magnetic touch and visual touch are fused. 

2. Magnetic field data and visual touch data are fused, and heterogeneous data are combined to use convolutional neural networks (CNN) and recurrent neural networks (RNN) to achieve higher force estimation accuracy.

3. A series of experiments were conducted to verify the multimodal perception capability of the MagicGel sensor fusion method. MagicGel using this fusion method has force measurement, fast response time, and proximity perception capabilities.

\section{RELATED WORK}

\subsection{VBTS Design}

The fundamental principle of the VBTS is to utilize a camera to capture images of the surface in contact. Surface contact information is extracted by processing and analyzing the acquired images \cite{sun2022soft}. However, the changes that occur after the surface makes contact with an object—such as marker displacement \cite{ma2019dense} and texture alterations \cite{lu2024innovative}—exist in three-dimensional space. The camera compresses this three-dimensional information into a two-dimensional image, which inevitably results in information loss. Consequently, this loss can diminish the accuracy of force measurements. 

In order to improve the accuracy of force measurement in VBTS, \cite{zhang2018robot} uses a binocular system and \cite{zhang2022tac3d} sets up a virtual binocular device to reconstruct three-dimensional deformation. Although this method is relatively effective in improving the accuracy of force measurement. But at the same time, it will also pose difficulties and challenges to the miniaturization and integration of VBTS.
In addition, when designing VBTS, different markers are designed to obtain richer visual information, such as double-layer markers\cite{lin2020curvature}. However, the structure and manufacturing process of double-layer markers is complex, requiring more precise process. At the same time, in terms of optics, the multi-layer structure will introduce more complex diffusion and refraction distortion, increasing the complexity of the model. And \cite{li20233} designed a special structure to improve the sensor, and obtain higher force estimation accuracy through mutual verification of information. However, its special structure makes the force estimation less robust and not conducive to miniaturization.

\subsection{Multimodal perception}

VBTS is a tactile sensor capable of detecting and providing feedback on an object's surface texture and contact status. When this tactile sensor is integrated with a dexterous hand in practical scenarios, it is the primary feedback mechanism. To effectively navigate perception challenges in complex environments, there is a common practice of optimizing or altering the structure or materials based on the original design. In a study detailed in \cite{yu2020vision}, the author successfully implemented temperature perception using VBTS technology by incorporating temperature-responsive materials on the grating marking layer. Typically, robots rely on visual cameras for object positioning and manipulation tasks before contacting \cite{liu2024design}. However, in low-light conditions or instances of visual obstruction, this poses challenges for dexterous hand operations, particularly in contexts where collision avoidance is crucial. Adjusting control strategies through non-contact perception can also improve the efficiency of robots. Therefore, we have improved the visual-tactile sensor to achieve close-range perception based on magnetic field changes.

\begin{figure*}[t]
\centering 
\includegraphics[width=1\textwidth]{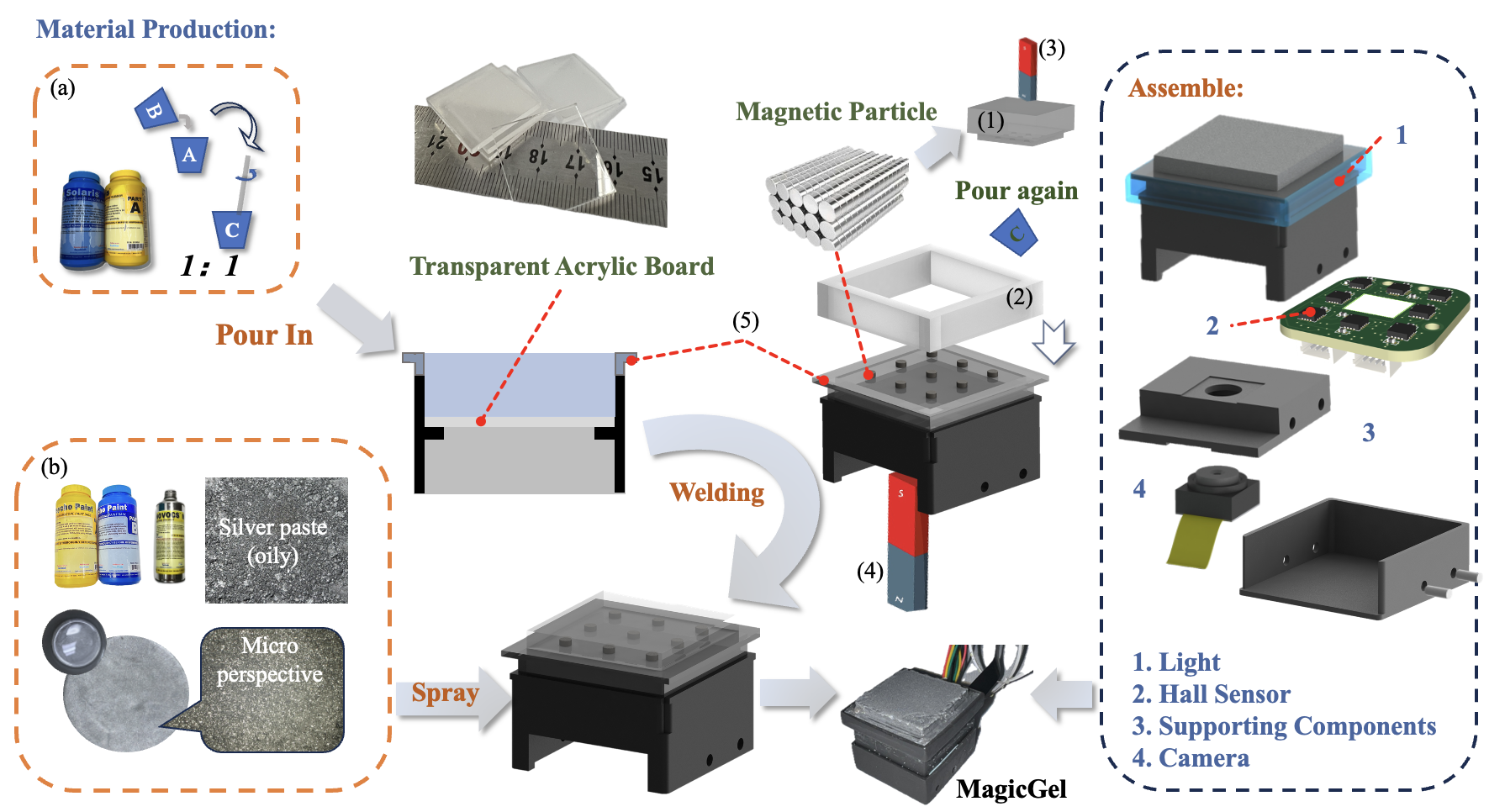} 
\caption{MagicGel manufacturing process flow chart. In the figure, (1) and (2) are molds 1 and 2, and (3) and (4) are magnets 1 and 2, (5) is transparent light source isolation housing. \textbf{(a)} Silicone material. \textbf{(b)} Surface spray material manufacturing process and spray quality (microscopic). The eight Hall sensors in Assemble 2 are evenly integrated into the circuit board.} 
\label{fig:2} 
\end{figure*}

\section{STRUCTURE AND MANUFACTURING}
\subsection{Structure of tactile sensor} 
The structure of the MagicGel is shown in Figure \ref{fig:1}. They are coating, magnetic particle marking point, elastomer, light strip, Hall sensor, and camera.

We divide the visual-tactile system into three components: the tactile module, the lighting module, and the camera module \cite{zhang2022hardware}. 

The tactile module primarily consists of a coating, marking points, an elastomer, and a support structure. The shape and size of the elastomer significantly influence the perception performance of the VBTS \cite{zhang2023improving}. During the experimental verification process, we confirmed that the force range was between 0 and 1 N, and the dimensions of the elastomer used were 25*25*11mm. The upper portion of the tactile elastomer was maintained in a low-constrained state. Tiny rigid magnetic particles served as marking points in the visual image.The chemical formula of the small magnet is Nd$_2$Fe$_{14}$B. The small magnetic particles used have magnetic force on each other, so 3x3 particles are placed 1mm below the surface coating. 
The distance between the marks is 5 mm. 

Lighting Module: A small light strip was employed to illuminate the internal environment and enhance the visibility of the coating contact image. We utilized small magnetic particles for setting the marking points. Although the diameter of the particles is only 1 mm, their thickness can still affect the quality of the received image when the light source is positioned internally. To mitigate this issue, a scattered linear light source was installed 1-2 mm below the plane where the small magnetic particles are located on the side. This arrangement ensures uniform brightness across the marking points. 

Camera Module: The camera module (WN-L2112.K314L) has a maximum receiving frame rate of 30 frames per second. In practical applications, this frame rate can be consistently achieved, meeting the requirements for standard VBTSs available on the market. 

The magnetic sensing component is primarily divided into a magnetic field-generating device (magnetic particle marker) and a magnetic receiving device (Hall sensor: MLX90393). We positioned eight Hall sensors evenly around the camera to detect the magnetic field generated by the entire array of magnetic particles. The overall size of MagicGel is 31*31*27mm.

\subsection{Manufacturing process}

In the manufacturing process of the MagicGel, it is essential to consider not only the optical path design, imaging distance, and coating but also the placement of the Hall sensor and the magnetic field generated by the initial magnetic particles.

The silicone material used in the VBTS significantly influences the force-sensing range and accuracy of the sensor. To meet our specifications, we selected Smooth-On's Solaris® silicone. Since the magnetic particles must be embedded within the elastomer, we employ a two-step casting method during the manufacturing process. Initially, we secure a transparent acrylic plate to the sensor housing using Kafuter® sealant to ensure a strong bond between the acrylic plate and the sensor body. The tactile component of the tactile sensor, which incorporates visual-magnetic fusion, remains in an unconstrained state and must be fixed and sealed onto the housing using transparent light source isolation housing, with a release agent applied internally. The transparent elastomer, made from Solaris® silicone, is chosen for its excellent light transmission properties. We mix the silicone components in a 1A:1B ratio by weight, stirring in one direction to minimize bubble formation. After confirming that the silicone is thoroughly mixed, we utilize vacuum equipment to eliminate any remaining bubbles. The prepared mixed solution is then injected onto the designated horizontal plane and placed in a vacuum environment to facilitate solidification.

During the process of applying magnetic particle markings, an attraction is generated between the magnetic particles. Consequently, the marking points may not be accurately positioned according to the pre-designed layout. If a groove is pre-formed in the first layer of silicone-based on the designated marking position, we have observed that certain defects can arise in the final elastomer, adversely affecting imaging quality. To ensure the correct positioning of the magnetic particles, we utilized the printing mold 2, as illustrated in Figure \ref{fig:2}. A strong magnet 1 is positioned beneath mold 2, and a groove is incorporated above mold 2 to securely place and hold the magnetic particles. The strong attraction of the magnet below ensures that the magnetic particles do not interfere with one another due to this attraction. Subsequently, mold 2, with the fixed magnetic particles, is inverted and placed onto the previously solidified elastomer. Simultaneously, the corresponding strong magnet 2 is positioned beneath the elastomer. Following this, strong magnet 1 and mold 2 with the groove are removed vertically from the top in succession. At this point, the strong magnet below exerts a force between the magnetic particles and the elastomer. The friction generated counteracts the attraction between the magnetic particles, thereby stabilizing the position of the small magnet. The aforementioned silicone is then injected to submerge 1 mm above the small magnet, which corresponds to the upper end of mold 1.

After solidification, the silicone material is utilized for surface spraying. Since the sensor component is in direct contact with external objects, it is essential to consider not only the wear resistance of the coating but also the adhesion between the coating and the gel-deformable body. To fulfill this requirement, we selected Smooth-On's Psycho Paint® silicone and NOVOCS™ Matte solvent. The addition of the silicone solvent to the mixed Psycho Paint® silicone imparts the characteristics of a water-like liquid. When choosing reflective pigments for MagicGel surface spray materials, silver paste aluminum paint is an excellent option, as it provides a strong reflective effect. To prepare the coating, add the silver paste aluminum paint to the silicone base in a specific proportion, then stir for five minutes. Pour the upper coating silicone into a new container and repeat this process 2-3 times to eliminate any undissolved silver paste particles. This results in a saturated solution of silver paste, which can then be evenly sprayed onto the surface using a spray gun.

\section{METHOD}

\begin{figure}[t]
\centering 
\includegraphics[width=0.5\textwidth]{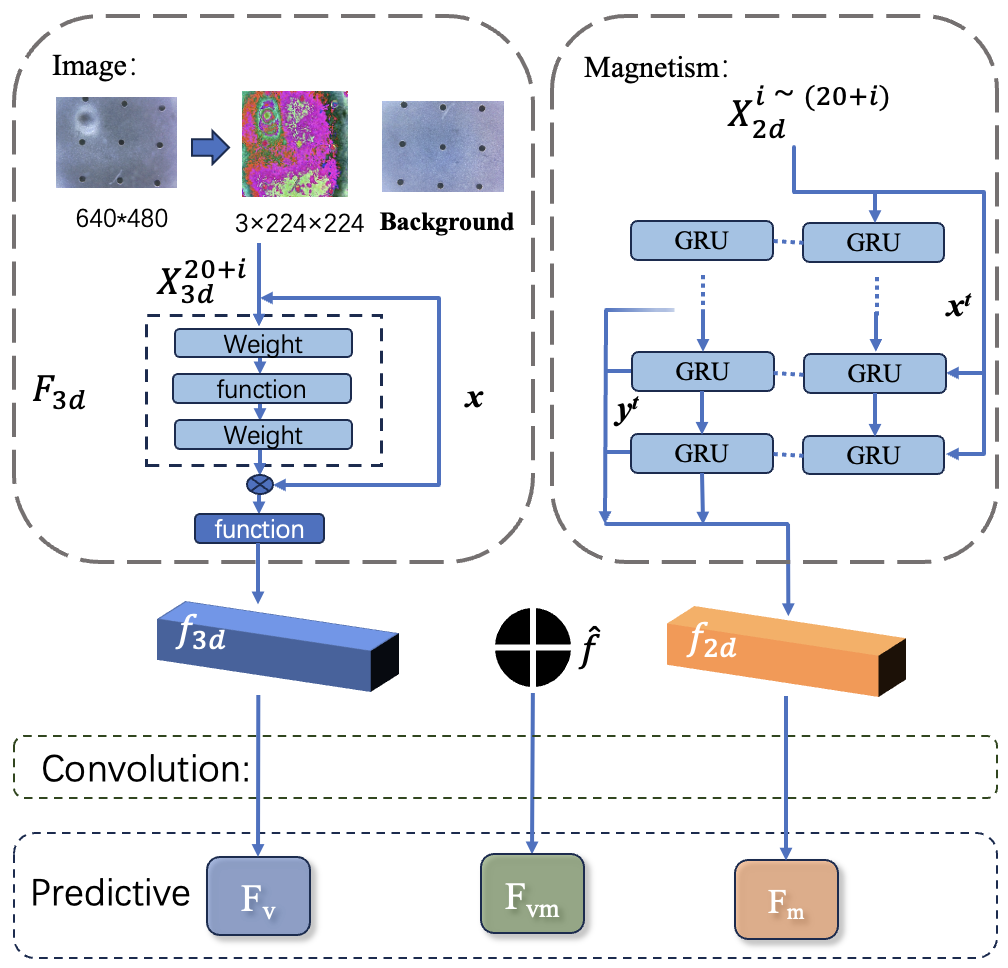} 
\caption{Schematic diagram of the network model for force measurement using VBTS. The RGB three-channel information of the tactile image and the 2D information of the magnetic tactile image are spliced and integrated to achieve the fusion of visual and magnetic information.} 
\label{fig:3} 
\end{figure}

This paper proposes the viewpoint of integrating magnetic information with tactile images for regression prediction. We utilize recurrent neural networks to process the feature vectors obtained from magnetic information and combine them with feature vectors processed from images using convolutional neural networks. The feature vectors from both networks are concatenated to predict the normal force on the sensor surface, as shown in figure \ref{fig:3}. The well-trained model achieved the best root-mean-square error of 0.0497 N. Our focus is on validating the algorithm's feasibility by concentrating on the normal force. The results demonstrate a significant improvement in accuracy compared to using images alone, thereby confirming the effectiveness of the visual-magnetic integration approach.

Network Architecture Overview: The model takes aligned images and magnetic field data as input.  The collected images, after undergoing preprocessing such as compression and normalization, are fed into a custom convolutional network to extract image feature vectors. The magnetic data from 20 consecutive moments are processed through ready-made Gated-Recurrent-Units (GRU), using the hidden state from the final time step as the feature vector. The feature vectors from both modalities are directly concatenated and then passed through a fully connected layer for regression prediction. The mathematical expression of the model is as follows:

\setlength{\abovedisplayskip}{2pt}
\begin{flalign}
    &&
    {\hat{y}} = F_f{(X^{n+i}_{3d}},{X^{i \sim (n+i)}_{2d}}) \in \mathbb{R}^1 
    &&
    \\[10pt]
    &&
    {X^{20+i}_{3d} \in \mathbb{R}^{3 \times H  \times W }}, {X^{i \sim (20+i)}\in\mathbb{R}^{n \times m}}
    &&
\end{flalign}

Where $\hat{y}$ represents the predicted normal force for the $n+i$ frame. Specifically, $X^{n+i}_{3d}$ denotes the image at the $n+i$ time step, while $X^{i \sim (n+i)}_{2d}$ refers to the magnetic data from time $i$ to $n+i$. Here, The parameter $n$ represents the number of continuous observation moments. $m$ represents the length of the one-dimensional magnetic data. In this study, eight Hall elements are utilized, each measuring the magnetic field strength in the $x$, $y$, and $z$ directions, resulting in $m$ being equal to 24. Additionally,$H$ and $W$ represent the height and width of the standardized image, respectively, with $H = W = 224$ in this paper.

\begin{figure}[t]
\centering 
\includegraphics[width=0.5\textwidth]{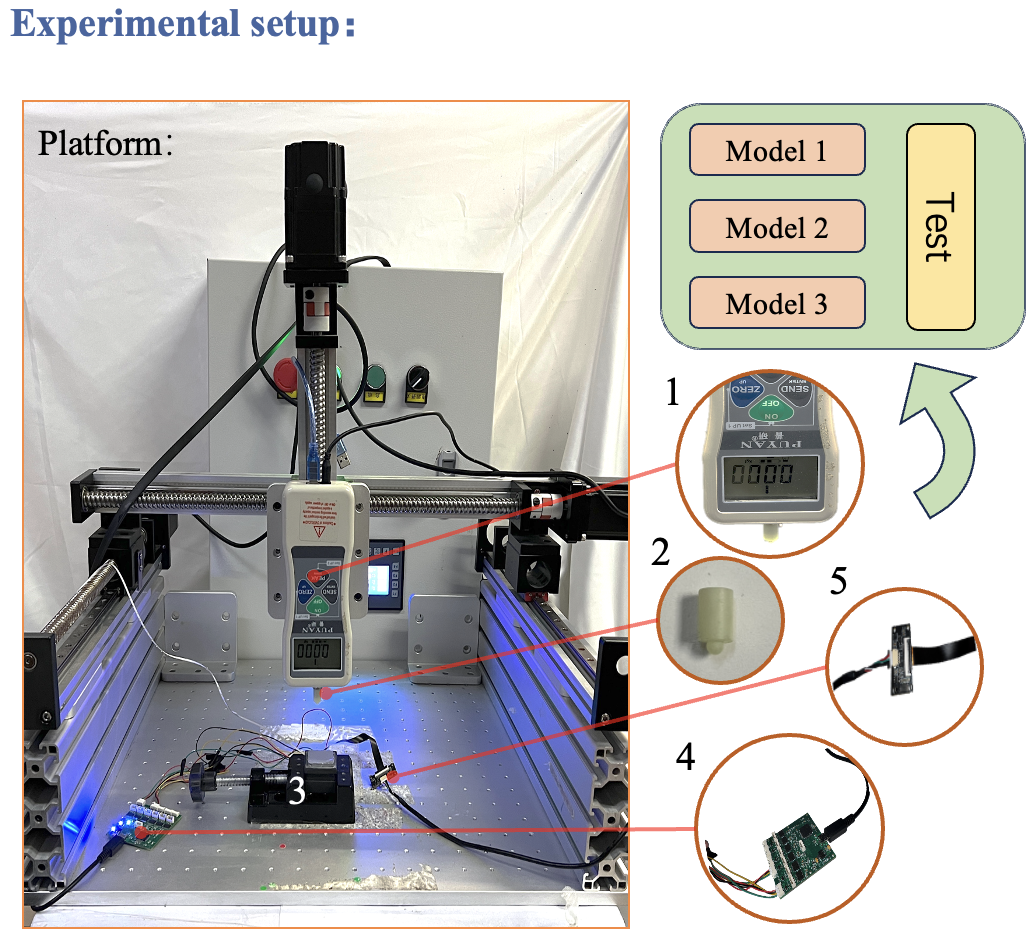} 
\caption{\textbf{(1)} Dynamometer(Standard force, Move horizontally, press vertically). \textbf{(2)} Pressure head. \textbf{(3)} MagicGel. \textbf{(4)} Magnetic Data Module. \textbf{(5)} Visual data receiving module. Model 1, model 2 and model 3 are models trained on three different sets of data. 
}
\label{fig:5} 
\end{figure}

\subsection{Experimental method for normal force estimation}

\subsubsection{Task Definition}

The prediction of normal force is considered a regression problem. Utilizing three-dimensional images and magnetic data obtained from sensors, the network infers the normal force for each frame.

\subsubsection{Normal force estimation network}

The image data is processed into a one-dimensional feature vector $f_{3d}$ using the custom model $F_{3d}$, while the magnetic data is transformed into a one-dimensional feature vector $f_{2d}$ using the $F_{2d}$ (GRU) model.

\begin{flalign}
    &&
    \label{f4}
    f_{3d} = F_{3d}(X^{n+i}_{3d}), f_{2d} = F_{2d}(X^{i \sim (n+i)}_{2d})
    &&
\end{flalign}

Since magnetic information can be easily influenced by environmental noise, which affects normal force estimation, we utilize magnetic information from continuous moments, aiming to learn effective changes in magnetic data based on sensor deformation from multi-moment data. These features are subsequently fed into the fully connected layer $ F_{c} $ to predict the normal force  $\hat{y}$ .

\begin{flalign}
    \label{f6}
    &&
    {\hat{y}} = f_{2d} \oplus f_{3d}, {\hat{y}}=F_{c}({\hat{f}})
    &&
\end{flalign}

\subsubsection{Normal force estimation method and experimental setup}

To determine whether supplementing image information with magnetic can enhance the accuracy of force prediction, we conducted a comparative experiment on force prediction accuracy. This section introduces the normal force measurement platform and the normal force dataset.

\begin{itemize}
\item Platform building. The sensor is mounted on a three-degree-of-freedom mobile calibration platform. The force gauge equipped with indenters is systematically maneuvered to collect standard force, image, and magnetic data. The pressure head is a 3D-printed part made of resin material, which greatly reduces the disturbance to the magnetic field. The experimental setup is shown in Figure \ref{fig:5}.
\item Dataset collection. The dataset was collected by applying Normal pressure with a dynamometer at 400 locations arranged in a 20x20 grid on the sensor's surface. This process collected 17,000 sample sets, with force measurements ranging from 0 to 1.0 N. The dataset was then divided into training, testing, and validation sets in an 8:1:1 ratio.
\item Network implementation details. Image data undergoes a module consisting of convolution, batch normalization, and activation, followed by flattening and fully connected layers to obtain feature vectors. The magnetic data input to the model is formatted as the most recent 20 moments of magnetic data. The experiments were conducted using the PyTorch framework and trained on an NVIDIA GTX 1650 GPU. We employed the Adam optimizer with varying learning rates, a weight decay of 0.0001, and utilized Smooth L1 Loss as the loss function, with a batch size of 64. Training is terminated if there is no improvement in the loss on the test set for three consecutive epochs. The model parameters are detailed in Table \uppercase\expandafter{\romannumeral1}.
\item Evaluation method. We utilized RMSE, MSE, and R² to assess the estimation error of the normal force and the accuracy of the model's regression.
\end{itemize}

\subsection{Fast response verification method}

On our constructed platform, we employed the same method to verify the response sensitivity across different modalities. We collected and compared the force measurement feedback rates using the three distinct modalities of MagicGel—single magnetic, single image, and image-magnetic fusion—including data collection time, data preprocessing time, and model inference time. We conducted comparative experiments with the Hall element baud rate set to 1,000,000 and the camera frame rate at 30 frames per second.

\subsection{Proximity Verification}

The magnetic provides the sensor with the ability to perform non-contact prediction, thereby enhancing the tactile sensor's capacity to detect and locate objects without physical contact. We conducted non-contact perception experiments using a variety of common electronic devices, including smartwatches, Bluetooth headsets, mobile phones, and computer mice. The MagicGel tactile sensor was mounted on the experimental platform, and the objects were gradually brought within 20 cm of the sensor to record changes in the magnetic data. 

\begin{table}[t]
\centering
\caption{Network Parameter Table}
\label{tab:1}
\begin{threeparttable}
\begin{tabular}{c|cc}
\hline
\centering
Network & \multicolumn{1}{c|}{CNN} & GRU \\ \hline
Input shape & \multicolumn{1}{c|}{3*224*224} & 20*24 \\ \hline
Model parameter & \multicolumn{1}{c|}{\begin{tabular}[c]{@{}c@{}}CBR1(3,16,7,2,3)\\ CBR2(16,32,5,2,2)\\ CBR3(32,64,3,2,1)\\ Flatten\\ FC1(64*28*28,512)\\ FC2(512,32)\end{tabular}} & \begin{tabular}[c]{@{}c@{}}in\_size 24\\ \\ num\_layers 3\\ hidden\_size 32\end{tabular} \\ \hline
Output shape & \multicolumn{1}{c|}{1*32} & 1*32 \\ \hline
Fc out & \multicolumn{2}{c}{(64,1)} \\ \hline
\end{tabular}

\begin{tablenotes}
\footnotesize
\item[1]CBR(3,16,7,2,3) indicates Conv convolution layer, BatchNorm2d normalization layer and ReLu activation layer.(kernel size = 7×7, input channels = 3, output channels = 16, Stride = 2, padding = 3)
\item[2] Flatten indicates compression into a one-dimensional tensor.
\item[3] FC(512,32) indicates a fully connected layer (input channels = 512, output channels = 32).

\end{tablenotes}           
\end{threeparttable}
\end{table}

\section{EXPERIMENTAL ANALYSIS}

\begin{figure*}[t]
\centering 
\includegraphics[width=1\textwidth]{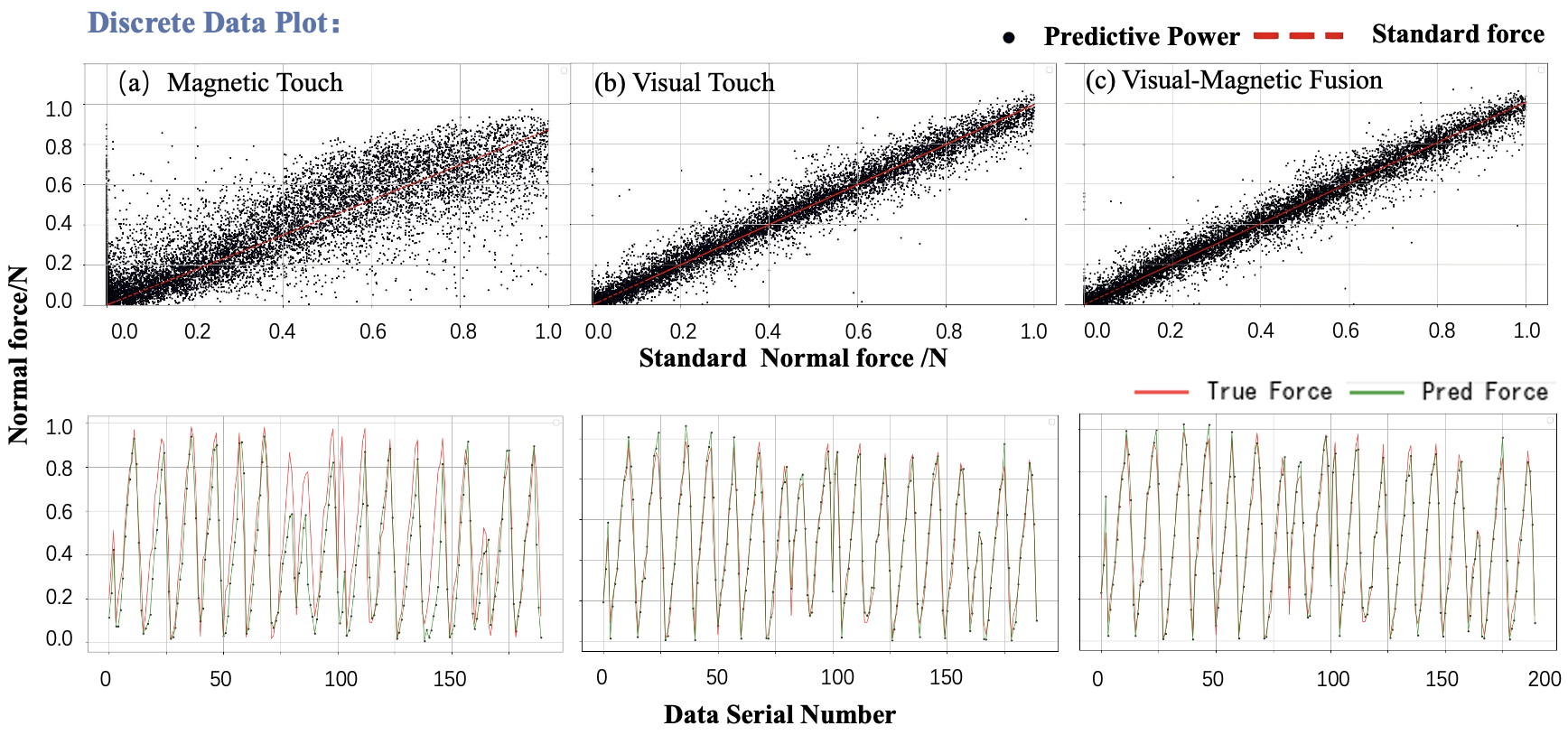} 
\caption{Force measurement results (where a, b, c are force/N, and d is the data number). \textbf{(a)} \textbf{(b)} \textbf{(c)}Magnetic tactile, Visual tactile force and Visual-magnetic fusion measurement results. \textbf{(d)} Visual-Magnetic force estimation verification result line chart. True value (red), estimated value (green).
The degree of dispersion of the points in the graph represents the degree of expressiveness estimation.
}
\label{fig:6} 
\end{figure*}

\subsection{Force measurement experimental results and analysis}
This paper proposes a structural improvement method to improve the force measurement accuracy of visual-tactile sensors. Therefore, the force measurement comparison needs to be made in the same visual-tactile perception environment. Therefore, in the experimental environment designed in this paper, the increase or decrease of magnetic field data is used as a variable to explore the influence of magnetic data on measurement accuracy.

To analyze the experimental data, we established three control groups. Group 1: Models were trained using magnetic data and force information. Group 2: Models were trained using visual images and force information. Group 3: Models were trained using visual images, magnetic data, and force information.

The experimental results demonstrate that our proposed method outperforms another algorithm. When the learning rate was set to 0.0001, a comparison of the model outputs from Groups 2 and 3 on the validation set revealed the impact of incorporating magnetic information into tactile imaging on force measurement accuracy. Simultaneously, the setup for Group 1 aims to demonstrate whether changes in the accuracy of force measurements, resulting from the incorporation of magnetic information in visual images, can be attributed to the intrinsic sensing characteristics of magnetic tactile perception. The RMSE for Group 1 was 0.1399, with an MSE of 0.0196. According to the results in Table \uppercase\expandafter{\romannumeral2}, the RMSE for the images was 0.0553, and the MSE was 0.0031. After integrating magnetic information with the two-dimensional visual images, the RMSE and MSE of the force measurement improved to 0.0497 and 0.0025, respectively, achieving a 10.1\% enhancement in the force measurement accuracy of the visual information. The results from Group 3 highlight the significant impact of incorporating magneto-tactile information on force measurement precision, demonstrating that the proposed visual-magnetic combination model(MagicGel) more accurately estimates normal force.

\begin{table}[h]
\setlength{\tabcolsep}{1.5mm}{
\centering
\caption{Force Accuracy Comparison Table}
\label{tab:}
\begin{tabular}{ccc|ccc}
\hline
\cellcolor[HTML]{FFFFFF}{ method} & input & metric & Lr=0.001 & Lr=0.0001 & Lr=0.00001 \\ \hline
1.GRU & \begin{tabular}[c]{@{}c@{}}Magnetism \\ 2D\end{tabular} & \begin{tabular}[c]{@{}c@{}}MSE\\ RMSE\\ $R^2$\end{tabular} & \cellcolor[HTML]{FFFFFF}{\color[HTML]{666666} \begin{tabular}[c]{@{}c@{}}0.0129\\ 0.1138\\ 0.8461\end{tabular}} & \cellcolor[HTML]{FFFFFF}{\color[HTML]{666666} \begin{tabular}[c]{@{}c@{}}0.0196\\ 0.1399\\ 0.7705\end{tabular}} & \cellcolor[HTML]{FFFFFF}{\color[HTML]{666666} \begin{tabular}[c]{@{}c@{}}0.0248\\ 0.1575\\ 0.7187\end{tabular}} \\ \hline
2.CNN & Image 3D & \begin{tabular}[c]{@{}c@{}}MSE\\ RMSE\\ $R^2$\end{tabular} & \cellcolor[HTML]{FFFFFF}{\color[HTML]{666666} \begin{tabular}[c]{@{}c@{}}0.0032 \\ 0.0563\\ 0.9644\end{tabular}} & \cellcolor[HTML]{FFFFFF}{\color[HTML]{666666} \begin{tabular}[c]{@{}c@{}}0.0031\\ 0.0553\\ 0.9657\end{tabular}} & \cellcolor[HTML]{FFFFFF}{\color[HTML]{666666} \begin{tabular}[c]{@{}c@{}}0.0038\\ 0.0618\\ 0.9571\end{tabular}} \\ \hline
\begin{tabular}[c]{@{}c@{}}3.GRU+\\ CNN\end{tabular} & \begin{tabular}[c]{@{}c@{}}Magnetism \\ and Image \\ 2D+3D\end{tabular} & \begin{tabular}[c]{@{}c@{}}MSE\\ RMSE\\ $R^2$\end{tabular} & \cellcolor[HTML]{FFFFFF}{ \begin{tabular}[c]{@{}c@{}} \color[HTML]{666666}0.0027\\ 0.0518\\ \color[HTML]{666666}0.9684\end{tabular}} & \cellcolor[HTML]{FFFFFF}{\begin{tabular}[c]{@{}c@{}}\color[HTML]{666666}0.0025\\ 0.0497\\ \color[HTML]{666666}0.9709\end{tabular}} & \cellcolor[HTML]{FFFFFF}{\begin{tabular}[c]{@{}c@{}} \color[HTML]{666666}0.0025\\ 0.0500\\ \color[HTML]{666666}0.9706\end{tabular}} \\ \hline
\end{tabular}}
\end{table}

\subsection{Multimodal feedback mechanism}

VBTS that incorporate magnetic awareness offer a wide range of application possibilities. The integration of visual perception, visual-magnetic fusion, and magnetic perception can more effectively meet diverse environmental needs. In robotic operations, the high sensitivity of magneto-tactile perception provides significant advantages over purely visual sensors in terms of both perception and functionality. The response rates of the different modes are shown in Table \uppercase\expandafter{\romannumeral3}, indicating that the MagicGel delivers faster feedback and significantly reduces response time, which is particularly beneficial for closed-loop operations in dexterous robotic hands. In environments with substantial magnetic interference, magnetic particles function exclusively as markers. The two-dimensional displacement and image alterations of these markers are recorded through a visual-tactile sensing mechanism, which operates as a comprehensive VBTS. For tasks that necessitate precise force measurement, this integration of visual-tactile and magnetic data offers an effective solution, enhancing the perceptual capabilities of the tactile sensor and enabling adjustments tailored to specific application requirements.

\begin{table}[t]
    \centering
    \renewcommand{\arraystretch}{1.5} 
    \caption{Response time comparison}
    \label{tab:Time}
    \begin{threeparttable}
    \begin{tabular}{c|c|c|c}
    \hline
    Time/ms & Data Reception & Preprocessing & Model Inference \\ \hline
    Ci & \textbf{18.842} & \textbf{0.690} & 3.092 \\ \hline
    Image & 21.411 & 2.956 & \textbf{3.002} \\ \hline
    Ci\_image & \backslashbox{}{} & 3.670 & 4.028 \\ \hline
    \end{tabular}

\begin{tablenotes}
\footnotesize
\item[1]In the case of visual-magnetic data fusion, data heterogeneity will affect the response speed.

\end{tablenotes}
\end{threeparttable}
\end{table}

\subsection{Proximity perception}

Tactile sensors typically rely on physical contact, processing information based on changes detected upon contact. However,  proximity perception capabilities significantly enhance a sensor's ability to locate and interact with objects in its environment. Our experiments, conducted with common electronic products, demonstrate that proximity perception is particularly effective in detecting the approach of household electronic devices. 
There are also differences in the contact standards and magnetic fields of different objects. Therefore, we use the state at the time of initial contact as the standard, normalize the proximity perception data and the standard data, and then analyze the results.
The results are shown in Figure 5. This proximity approach can sense the proximity of objects, thereby improving the robot's ability to detect and respond before physical contact is made.

\section{CONCLUSIONS}

This study proposed a force estimation method that uses magnetic field data to improve the dimension of visual tactile data. MagicGel was created by integrating magnetic perception capabilities based on VBTS. In addition to high-precision and fast-response force estimation capabilities, MagicGel also realizes non-contact perception of objects.

However, MagicGel currently shows certain limitations: the interaction force between magnetic particles affects the setting of higher-density markers. At the same time, when in contact with objects with large magnetic field interference, force estimation cannot be achieved through the MagicGel method. In future work, we will focus on increasing the density of magnetic markers and establishing a MagicGel simulation model. 

In general, this study innovatively proposed and verified a hardware design method that combines magnetic fields to improve visual tactile sensing capabilities. It achieved multimodality of visual tactile sensors. This progress provides innovative ideas and solutions for strengthening the force perception ability of VBTS and improving dexterous hand manipulation.

\begin{figure}[t]
\centering 
\includegraphics[width=0.45\textwidth]{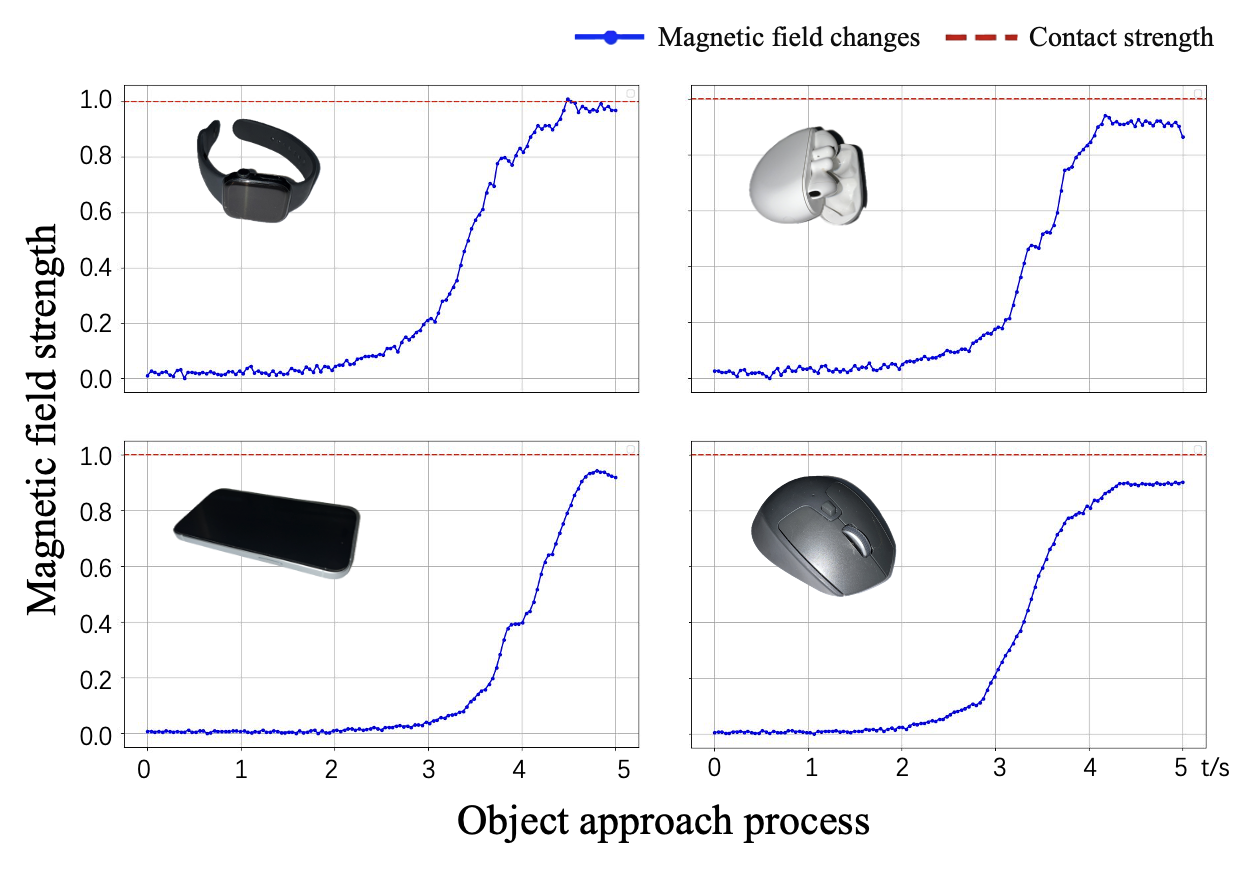} 
\caption{Proximity perception verification (watch, headset, mobile phone, mouse). The horizontal axis represents the time process of the object approaching MagicGel, and the vertical axis represents the degree of proximity (0-1).} 
\label{fig:4} 
\end{figure}

\section*{ACKNOWLEDGMENT}
This work was supported by the 2022-2024 Anhui Provincial Key Research Project on Natural Science of Higher Education Institutions and the Natural Science Research Project of Anhui Educational Committee (2022AH050306).

\addtolength{\textheight}{0cm}   





\bibliographystyle{IEEEtran}
\bibliography{reference}

\end{document}